\documentclass[letterpaper]{article} 
\usepackage{aaai2027}  
\usepackage[hyphens]{url}  
\usepackage{graphicx} 
\usepackage{natbib}  
\usepackage{caption} 
\usepackage{algorithm}

\usepackage{amsmath} 
\usepackage{amsfonts}
\usepackage{multirow}
\usepackage{algpseudocode}
\usepackage{graphicx}   
\usepackage{multirow}

\usepackage{threeparttable}
\usepackage{algorithm}
\usepackage{algpseudocode}
\usepackage{booktabs}
\usepackage{array}
\usepackage{amsmath, amssymb}

\usepackage{newfloat}
\usepackage{listings}
\DeclareCaptionStyle{ruled}{labelfont=normalfont,labelsep=colon,strut=off} 
\floatstyle{ruled}
\newfloat{listing}{tb}{lst}{}
\floatname{listing}{Listing}

\usepackage{booktabs}

\title{CoRS: Fast Neural Flexible Job-Shop Scheduling \\ via Composite Interval State Reuse{}}

\author{
    Sutong Jiang\textsuperscript{\rm 1},
    Peng Zhao\textsuperscript{\rm 1},
    Xuan Wu\textsuperscript{\rm 1},
    Yan Wang\textsuperscript{\rm 1},
    Jiaqi Wang\textsuperscript{\rm 2},
    Rui Zhang\textsuperscript{\rm 1}\corresponding,
    Yuan Jiang\textsuperscript{\rm 3}\corresponding,
    You Zhou\textsuperscript{\rm 1}\corresponding,
    Di Wang\textsuperscript{\rm 4}
}

\affiliations{
    \textsuperscript{\rm 1}College of Computer Science and Technology, Jilin University, China\\
    \textsuperscript{\rm 2}College of Software, Jilin University, China\\
    \textsuperscript{\rm 3}College of Computing and Data Science, Nanyang Technological University, Singapore\\
    \textsuperscript{\rm 4}Joint NTU-UBC Research Center of Excellence in Active Living for the Elderly, Nanyang Technological University, Singapore\\
  
}

\begin{document}

\maketitle

\begin{abstract}

Neural solvers for the Flexible Job-Shop Scheduling Problem~(FJSP) typically refresh graph representations and update feasibility after every operation--machine decision. However, only feasibility must remain exact at every step, whereas graph encoding is substantially more
expensive and can potentially be refreshed less frequently. We present CoRS, which decouples them: an interval-level graph representation is reused across multiple sequential decisions, while a stepwise feasibility state is updated after every action. This two-timescale design amortizes graph encoding without parallelizing action generation or relaxing scheduling constraints.  CoRS further develops a feasibility-preserving composite-interval decoding scheme with dynamic constraint tracking and boundary state synchronization, which reduces representation refresh frequency while maintaining continuous feasible schedule construction. Furthermore, to alleviate representation staleness caused by interval-level state reuse, CoRS introduces enhanced static features as a compensation mechanism. By providing more comprehensive structural information of the scheduling problem, these features improve the model's perception of overall problem characteristics. Extensive experiments on synthetic and multiple public FJSP benchmarks demonstrate that CoRS significantly reduces inference time while maintaining competitive solution quality, achieving favorable speed–quality trade-offs under tight wall-clock budgets and low-latency scenarios.
\end{abstract}


\section{Introduction}
With the advent of the Industry 4.0 era, the rapid development of information technologies such as the Internet of Things~(IoT)~\citep{li2015internet} and artificial intelligence~(AI)~\citep{mao2019opportunities} has accelerated the digitalization of manufacturing systems, making cloud manufacturing an important mechanism for modern manufacturing~\citep{bo2010cloud,ren2017cloud}. As a core issue in manufacturing systems~\citep{liu2015solving}, scheduling problems require efficient decision-making based on cloud data. With the increasing scale and complexity of manufacturing systems, the response speed of scheduling decisions has become a critical challenge in cloud manufacturing environments~\citep{kovacova2021sustainable,zhang2025dynamic}.

As a classical NP-hard Combinatorial Optimization Problem~(COP)~\citep{xie2019review}, the Flexible Job-Shop Scheduling Problem~(FJSP) has become a modeling prototype for diverse applications, including medical resource scheduling~\citep{abderrabi2021flexible}, semiconductor manufacturing~\citep{ghaedy2024simulation}, and aero-engine production~\citep{zhou2019multi}, owing to its enhanced flexibility~\citep{chaudhry2016research}. In practical manufacturing systems, scheduling systems are required to respond rapidly to changes in production states within limited time. Consequently, efficiently solving the FJSP has become a central challenge~\citep{akbaripour2018cloud,xu2025hybrid}.

Given the significance of FJSP, numerous methods have been proposed, including exact algorithms~\citep{tutumlu2023mip}, Priority Dispatching Rules~(PDRs)~\citep{branke2015hyper}, and metaheuristics~\citep{xu2025hybrid}. However, these methods suffer from limitations in solution quality, generalization, and computational efficiency~\citep{xiao2025improving,han2021deep,zhang2023deepmag}, making them inadequate to meet the requirements of cloud manufacturing. Recently, deep learning has introduced new methods for solving FJSP. Neural combinatorial optimization methods show strong modeling and decision-making capabilities for scheduling problems~\citep{wang2025expectation,liao2025bopo,smit2025neural}. Most studies formulate the scheduling process as a Markov Decision Process~(MDP), adopt autoregressive~(AR) decision-making, and integrate reinforcement learning for end-to-end optimization to obtain high-quality solutions~\citep{yuan2024solving,zhao2024actor}.

However, most existing neural FJSP solvers primarily focus on state representation learning or backbone architecture design~\citep{xu2024deep}, while implicitly assuming that the complete graph representation must be refreshed after every scheduling decision. In fact, the global state usually undergoes only limited changes between consecutive decision steps (see Appendix A for analysis). So existing methods  perform complete feature encoding for highly similar states, resulting in substantial  computation and significantly degrading solution efficiency.

Motivated by this observation, we question this assumption and explore a scheduling approach that reduces  state updates while preserving dynamic feasibility. We propose the Composite Regressive (CR) mechanism, which redefines the state update process in neural FJSP solving. Unlike existing methods that implicitly assume the complete graph representation must be refreshed after every scheduling decision, CR introduces representation-refresh frequency as an underexplored design dimension and formulates a two-timescale state evolution mechanism. Specifically, CR factorizes the scheduling state into two coupled components with different update frequencies: an interval-level representation state and a stepwise feasibility state. The interval-level representation state is refreshed only at composite-interval boundaries, while the stepwise feasibility state maintains dynamic constraints and ensures schedule feasibility by progressively updating the feasible action set.

Based on this mechanism, we further develop the \textbf{Co}mposite \textbf{R}egressive \textbf{S}cheduling~(CoRS) model for solving FJSP. Unlike blockwise or parallel autoregressive methods, CoRS
does not reduce the number of action-generation steps; it
reduces the number of representation-refresh steps. Unlike
naive representation caching, CoRS explicitly tracks the
evolving feasibility state and periodically reconstructs the
full scheduling state. By reusing interval-level graph representations across consecutive scheduling decisions, CoRS reduces the number of graph encoding operations from $T$ to $\lceil T/X \rceil$, while maintaining scheduling feasibility through the progressive update of the stepwise feasibility state.

However, reducing the frequency of state refreshes may cause the interval-level representations to fail to timely capture certain dynamic changes, thereby affecting the quality of action ranking. To alleviate this issue, CoRS introduces enhanced static features as a compensation mechanism. By providing more comprehensive structural information of the scheduling problem, these features improve the model's perception of global scheduling characteristics and enhance the effectiveness of interval representations under low-frequency updates. Meanwhile, state synchronization at interval boundaries further limits the accumulation of representation staleness, ensuring that its impact is mainly restricted to the ranking among feasible actions rather than compromising scheduling feasibility.

Specifically, we first propose a two-timescale state evolution mechanism that decouples the scheduling state into an interval-level representation state and a stepwise feasibility state. The interval-level representation state adopts composite intervals as the state refresh units rather than traditional decision steps, and is updated only at composite-interval boundaries. By reusing interval-level state representations across consecutive scheduling decisions, this mechanism reduces  encoding of complete graph representations and alleviates the inference overhead caused by frequent state updates. Meanwhile, the stepwise feasibility state continuously tracks dynamic constraint changes, ensuring the feasibility of sequential schedule construction. Furthermore, we introduce enhanced static features as a compensation mechanism for interval-level state reuse. By providing more comprehensive structural information of the scheduling problem, these features improve the model's perception of overall problem characteristics and mitigate representation staleness caused by low-frequency state refreshing. Extensive experiments conducted against four state-of-the-art methods~\citep{wang2023flexible,zhao2025dual,song2022flexible,xiao2026resched} and four well-known PDRs demonstrate that CoRS can generate competitive solutions while significantly improving inference speed.

The main contributions of this study are summarized as follows:
\begin{itemize}
    \item We expose representation-refresh frequency as an
underexplored design axis in neural FJSP solving and
formulate a two-timescale state evolution that decouples
interval-level graph representations from stepwise
feasibility updates.
    
    \item We develop a feasibility-preserving composite-interval
decoding scheme with dynamic constraint tracking and
boundary state synchronization. It reduces graph-encoder
invocations from $T$ to $\lceil T/X \rceil$ while retaining
sequential and feasible schedule construction.
    
    \item We provide same-backbone and interaction studies that
isolate the effect of representation reuse and show that
enhanced static features increasingly compensate for
representation staleness as the interval length grows.
Fixed-budget comparisons further demonstrate the practical
quality--latency trade-off.

\end{itemize}

\section{Related Work}

Traditional methods for solving FJSP mainly include exact methods, PDRs, and metaheuristic algorithms. Exact methods construct FJSP as a mathematical model and explore the entire solution space to find the optimal solution, such as branch and bound~\citep{brucker1994branch}, integer programming~\citep{meng2020mixed}, and constraint programming~\citep{muller2022algorithm}. PDRs directly guide the scheduling decision process using manually designed priority rules, such as the Shortest Processing Time~(SPT) rule~\citep{chen2013flexible} and the Most WorK Remaining~(MWKR) rule~\citep{branke2015hyper,sels2012comparison}. Metaheuristic algorithms iteratively explore the solution space through carefully designed search strategies to obtain high-quality solutions, such as genetic algorithms~\citep{gao2019review} and particle swarm optimization~\citep{nouiri2018effective,zhang2009effective}. However, exact methods struggle to handle large-scale instances~\citep{demir2013evaluation,zhang2023deepmag}. PDRs based on fixed heuristics fail to capture complex scheduling patterns, resulting in low solution quality~\citep{chen2013flexible}. Metaheuristic algorithms require extensive iterative searches, incurring high computational costs~\citep{caldeira2019solving}. Furthermore, traditional methods typically solve each unseen instance from scratch, which limits their generalization ability.

In recent years, deep learning-based neural solvers have received increasing attention for solving FJSP.  To enhance the expressive power for complex scheduling problems, existing neural FJSP solvers often incorporate graph neural networks~\citep{zeng2022deep} or Transformer-based architectures~\citep{xu2024deep} to model operation--machine interactions and precedence dependencies.Very recently,ReSched~\citep{xiao2026resched} explored simplified state representations for Transformer-based FJSP solving and reported promising results.
Parallel autoregressive methods, such as PARCO~\citep{berto2026parco} and MACSIM~\citep{luttmann2025multi}, alleviate the sequential dependency of conventional autoregressive approaches by predicting multiple actions within a single decoding stage. These methods mainly accelerate scheduling from the perspective of action generation. However, existing solvers typically still require the complete graph-state representation to be refreshed after each decision. Such frequent representation updates and environment interactions introduce substantial computational overhead, significantly increasing inference time. As the scale of FJSP grows, inference time increases linearly with the number of time steps~\citep{demir2013evaluation}, thereby exposing significant efficiency bottlenecks in large-scale or real-time scheduling scenarios.

Recently, some studies have explored fast solving approaches based on fixed state representations, such as non-autoregressive~(NAR) methods.  NAR methods have been widely applied to the traveling salesman problem~\citep{dat2025neufaco} and the vehicle routing problem~\citep{xiao2024distilling}. However, NAR models generally suffer from degraded solution quality~\citep{joshi2019efficient}. This limitation mainly arises because NAR models make decisions based on static state representations, making it difficult to capture the dynamic changes in scheduling states caused by sequential decisions. As a result, scheduling solutions can be generated efficiently, but they often deviate from high-quality solutions~\citep{xiao2024reinforcement}. In FJSP, the relationships among nodes are highly complex and the scheduling state changes continuously with decisions, making it difficult to complete all decisions under a fixed state representation. Consequently, to our knowledge, fully non-autoregressive neural construction methods that directly generate complete feasible FJSP schedules remain underexplored; existing fast-learning or two-stage approaches either decompose the problem, rely on solver-assisted post-processing, or still construct schedules sequentially.


Overall, existing neural FJSP solvers mainly focus on state representation learning and network architecture design, while representation-refresh frequency, as an important design dimension affecting inference efficiency, remains largely unexplored. Although fixed state representation methods can reduce update overhead, they struggle to adapt to dynamically changing feasibility constraints in FJSP. To address this problem, we propose the CoRS, which decouples interval-level representation reuse and stepwise feasibility updates through a two-timescale state evolution design, reducing state refresh overhead while maintaining dynamic feasibility during scheduling. This makes CoRS complementary to recent representation- and backbone-level advances in neural FJSP solving.

\section{Preliminaries}
This section introduces the CR mechanism and presents the formulation of the FJSP.

\textbf{Composite regressive mechanism.} The CR mechanism adopts composite intervals as the update units for graph-state representations, rather than re-encoding the complete graph representation after each scheduling decision. Within each composite interval, CoRS reuses the interval-level graph-state representation and only updates the stepwise feasibility state to track dynamically changing constraints during scheduling. At composite-interval boundaries, the interval-level graph-state representation is synchronously reconstructed, reducing redundant graph encoding overhead and preventing outdated information from accumulating due to infrequent representation updates. Through this two-timescale state evolution mechanism, CoRS reduces graph-state refresh frequency and inference overhead while ensuring that action selection satisfies dynamic constraints through stepwise feasibility updates, thereby achieving a better efficiency--quality trade-off.

Under the CR mechanism, the entire FJSP scheduling process is divided into multiple consecutive, non-overlapping composite intervals $B$. $B^\tau$ denotes the $\tau$-th composite interval, which contains $X$ time steps and is defined by:
\begin{equation}
B^\tau=\{t\mid(\tau-1)X+1\leq t\leq \min(\tau X,t_{\mathrm{final}})\}
\end{equation}
where $\tau=1,2,\dots,\tau_{\max}$, $t_{\text{final}}$ denotes the final time step of the entire scheduling process, and $\tau_{\max}$ represents the total number of composite intervals.

\textbf{Flexible job-shop scheduling problem.} The FJSP is defined as follows. Given a set of $n$ jobs $J=\{J_1,J_2,\dots,J_n\}$ and a set of $m$ machines $M=\{M_1,M_2,\dots,M_m\}$, each job $J_i$ consists of $n_i$ operations $O_i=\{O_{i1},O_{i2},\dots,O_{in_i}\}$. Each operation $O_{ij}$ has a compatible machine set $M_{ij}\subseteq M$ and can be processed on any machine $M_k\in M_{ij}$ with processing time $p_{ijk}$, where the completion time is denoted by $C_{ij}$. The objective is to minimize the maximum completion time $C_{\max}=\max\{C_{in_i}\mid i=1,2,\dots,n\}$. The scheduling process must satisfy the following constraints: the operations of each job must follow their precedence constraints, and each machine can process at most one operation at a time with non-preemptive processing.

\begin{figure*}[htbp]
  \centering
  \includegraphics[width=0.9\linewidth]{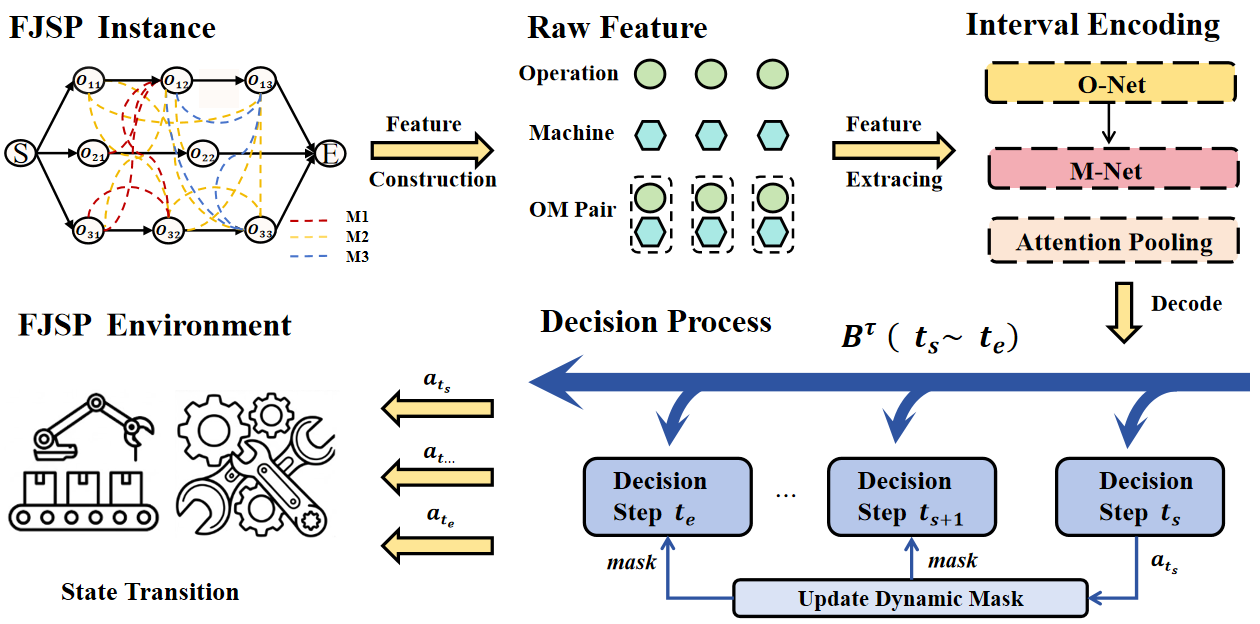}
  \caption{The architecture of CoRS.}
  \label{fig:architecture}
\end{figure*}

\section{Methodology}

This section introduces the CoRS model, the scheduling model based on the CR mechanism, which achieves efficient inference for the FJSP. Figure~\ref{fig:architecture} illustrates the workflow of CoRS. Through a specially designed interval encoding network and decoding algorithm, CoRS generates high-quality scheduling policies while significantly reducing inference time.  In the following, we sequentially present the MDP, the feature encoding scheme, the interval encoding network, the composite-interval decoding algorithm, and the reinforcement learning algorithm.

\subsection{MDP Formulation}

Unlike the MDP used in traditional reinforcement learning, under the CR mechanism, CoRS formulates a tailored MDP for FJSP by considering the characteristics of reusing graph representations within each composite interval and updating the graph representations across intervals. The MDP is defined as follows.

\textbf{State.} Within the composite interval $B^{\tau}$, CoRS adopts a two-timescale state evolution mechanism, which decomposes the scheduling state into an interval-level representation state $s^{\tau}$ and a stepwise feasibility state $s_t^{\mathrm{feas}}$ with different update frequencies. Specifically, $s^{\tau}$ is defined as $s^{\tau} = \{h_{ij},\, h_k,\, h_{(ij,k)}\}$, where $h_{ij}$, $h_k$, and $h_{(ij,k)}$ denote the features of operation $O_{ij}$, machine $M_k$, and operation--machine $(O_{ij}, M_k)$ pair, respectively. Meanwhile, $s_t^{\mathrm{feas}}$ records the step-wise updated feasibility 
information, including candidate operations, operation--machine 
feasibility masks, and machine 
release times. Different from the $s^{\tau}$, $s_t^{\mathrm{feas}}$ is only used to determine the feasible 
action space and does not participate in state feature encoding or 
reconstruction within the composite interval.

\textbf{Action.} For any time step $t$, the action space $A_t$ consists of all feasible OM pairs at the current scheduling state. The model selects one OM pair from $A_t$ as the action $a_t$.     

\textbf{State Transition.} Under the CR mechanism, state transitions follow a two-timescale update process. For the interval-level representation state, the state is synchronized only at composite-interval boundaries. The next state is defined as $s^{\tau+1} = f_{\mathrm{recon}}\!\left(s^{\tau}, \{a_t\}_{t \in B^{\tau}}\right)$, where $f_{\mathrm{recon}}(\cdot)$ denotes the vectorized state reconstruction function based on the action sequence within the composite interval. Meanwhile,  $s_t^{\mathrm{feas}}$ is updated after each operation--machine assignment to track dynamic constraint changes and maintain the feasible action space at each decision step.

\textbf{Reward.} For any operation $O_{ij}$ at time step $t$, its estimated completion time is denoted as $C_{\mathrm{LB}}(O_{ij}, t)$. For scheduled operations, $C_{\mathrm{LB}}(O_{ij}, t)$ equals the actual completion time $C_{ij}$. For unscheduled operations, the estimated completion time of all initial operations is initialized as $C_{\mathrm{LB}}(O_{i0}) = 0$. The estimated completion time is recursively computed based on the completion time of the predecessor operation and the minimum processing time among candidate machines, given by $C_{\mathrm{LB}}(O_{ij}, t) = C_{\mathrm{LB}}(O_{i,j-1}, t) + \min_{k \in M_{ij}} p_{ijk}$. Accordingly, an estimate of the overall makespan at time step $t$ is given by $C_{\mathrm{LB}}(t) = \max_{1 \le i \le n,\; 1 \le j \le n_i} C_{\mathrm{LB}}(O_{ij}, t)$. During the scheduling process, the estimated completion times are updated as the agent executes actions. The reward is defined as the difference between the estimated makespans at two consecutive time steps, given by $r_t = C_{\mathrm{LB}}(t) - C_{\mathrm{LB}}(t+1)$. 

\subsection{Feature Encoding}

Unlike conventional methods that re-encode state representations at each decision step, CoRS reduces redundant encoding operations by reusing interval-level representation state within composite intervals. Therefore, the state representation obtained from a single encoding process needs to incorporate more instance-related static information to support effective decision making over subsequent multiple scheduling steps. Based on this consideration, we propose an enhanced static feature encoding scheme. Specifically, given an FJSP instance as input, we encode it into operation features $h_{ij} \in \mathbb{R}^{12}$, machine features $h_k \in \mathbb{R}^{10}$, and OM pair features $h_{(ij,k)} \in \mathbb{R}^{8}$. We enhance the static features by incorporating information such as the average processing time of successor operations and the total number of fastest operations for each machine, as detailed in Appendix B.

\subsection{Interval Encoding Network}
\begin{figure}[htbp]
  \centering
  \includegraphics[width=0.9\linewidth]{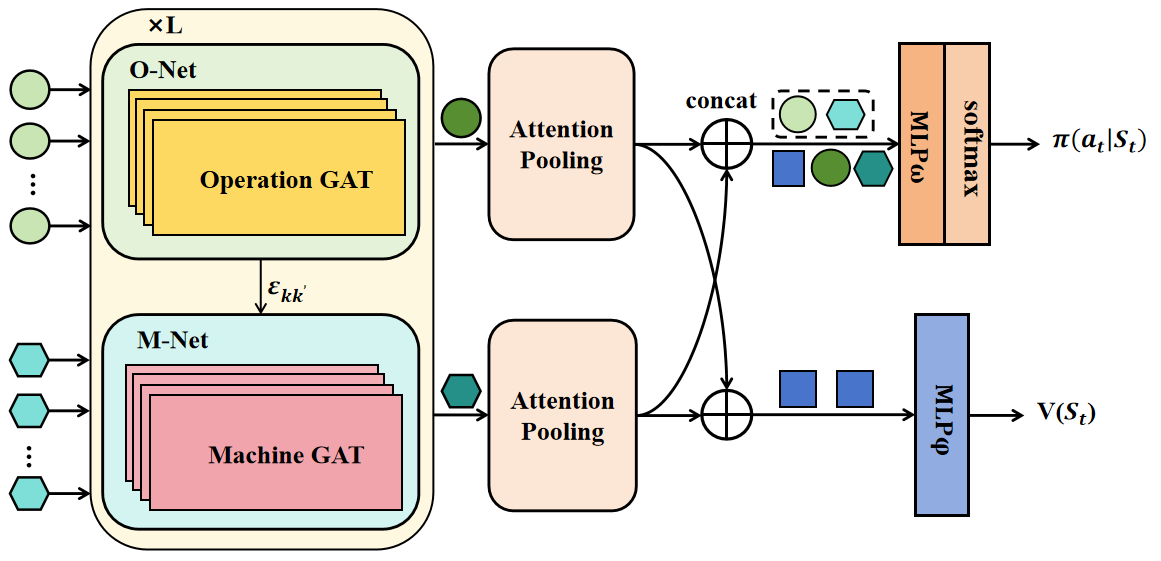}
  \caption{The network architecture of the CoRS.}
  \label{fig:network}
\end{figure}

Under the CR mechanism, the CoRS performs unified encoding of interval-level representation state features only at the initial step of each composite interval, reuses the resulting feature representations throughout subsequent consecutive decision steps, and re-executes global encoding after composite interval completion. As illustrated in Figure~\ref{fig:network}, the CoRS model employs a graph attention framework composed of an Operation Encoding Subnetwork~(O-Net) and a Machine Encoding Subnetwork~(M-Net) to model operation nodes and machine nodes, respectively.

\subsubsection{Operation Encoding Subnetwork}
O-Net models the operation states within the composite interval. In the $l \in L$-th multi-head graph attention layer, $h_{ij}$ is first linearly projected as $z_{ij} = W h_{ij}$, where $z_{ij} \in \mathbb{R}^{H d_h}$ and $W \in \{W', W_1 W_2\}$. $H$ denotes the number of attention heads and $d_h$ denotes the output dimension of each head. The selection of the projection matrix $W$ depends on the layer index. For the first layer, a full shared matrix $W' \in \mathbb{R}^{H d_h^{(1)}\times d_{in}^o}$ is used to map the input features into the representation space of all attention heads. For intermediate layers, a low-rank factorization $W_1 W_2$ is adopted, where 
$W_1 \in \mathbb{R}^{Hd_h^{(l)} \times d_{mid}}$ and $W_2 \in \mathbb{R}^{d_{mid} \times Hd_h^{(l-1)}}$ with $d_{mid} < H \times d_h^{(l-1)}$, which reduces the number of parameters while preserving the projection dimension.

The vector $z_{ij}$ is split into $H$ sub-vectors as $z_{ij}=[z_{ij}^{(1)} \| \cdots \| z_{ij}^{(H)}]$, where $[a \| b]$ denotes vector concatenation and $z_{ij}^{(h)} \in \mathbb{R}^{d_h}$. To mitigate the reduced head diversity caused by the shared projection, a learnable scaling parameter $\beta^h$ is applied to obtain $\tilde{z}_{ij}^{h}=\beta^{h}\odot z_{ij}^{(h)}$. In the $h$-th attention head, the attention coefficient between operation $O_{ij}$ and its neighboring operation $O_{ip}$ is computed as:
\begin{equation}
e_{ijp}=\mathrm{LeakyReLU}\!\left(a^{\top}[\tilde{z}_{ij}^{h}\|\tilde{z}_{ip}^{h}]\right), 
\quad p\in\{j-1,j,j+1\}
\end{equation}
where $a\in\mathbb{R}^{2d_h}$. The coefficients $e_{ijp}$ are normalized by a softmax function to obtain the attention weights $\alpha_{ijp}$. The output feature of this head is computed as:
\begin{equation}
\tilde{h}_{ij}^{h}=\sum_{p\in N_{ij}}\alpha_{ijp}\tilde{z}_{ip}^{h}
\end{equation}
where $N_{ij}=\{j-1,j,j+1\}$. After obtaining the operation features from all attention heads, the intermediate layers concatenate the multi-head outputs, while the final layer adopts mean aggregation. The unified operation node representation is computed as $h_{ij}^{\prime(\tau)}=\mathrm{ELU}\!\left(\frac{1}{H}\sum_{h=1}^{H}\tilde{h}_{ij}^{h}\right)$.

\subsubsection{Machine Encoding Subnetwork}

M-Net encodes machine node features to generate representations of machine states. In addition, M-Net introduces competition edge features. Specifically, $I_{kk'}$ denotes the set of unscheduled operations jointly competed by machines $M_k$ and $M_{k'}$. If $I_{kk'} \neq \emptyset$, the two machines are considered to have a competition relationship. Accordingly, the set of machines competing with $M_k$ is defined as $N_k=\{M_{k'} \mid I_{kk'} \neq \emptyset\}$. To quantify the competition intensity, we define the competition strength between them as the aggregation of the features of candidate operations in the candidate set $J_c$, given by $\epsilon_{kk'}=\sum_{O_{ij}\in I_{kk'} \cap J_c} h_{ij}$.

At the feature projection stage, we follow the same design as O-Net. $\tilde{z}_k$ and $\tilde{z}_{k'}$ denote the projected representations of $M_k$ and $M_{k'}$, and $\tilde{z}_{kk'}$ denotes the projected representation of their competition edge. The attention coefficient is computed as $u_{kk'}=\mathrm{LeakyReLU}(b_k^{\top}\tilde{z}_k^h+b_{k'}^{\top}\tilde{z}_{k'}^h+b_{kk'}^{\top}\tilde{z}_{kk'}^h)$, where $b_k,b_{k'},b_{kk'}\in\mathbb{R}^{d_h}$.  $u_{kk'}$ is normalized by softmax to obtain the attention weight $\alpha_{kk'}$, which is used to aggregate machine features and produce the single-head output $\tilde{h}_k$. After obtaining the outputs of all heads, M-Net adopts the same multi-head fusion strategy as O-Net to generate the final machine embedding $h_k^{\prime(\tau)}$.

\subsubsection{Attention Pooling}
After encoding operation and machine nodes with O-Net and M-Net, CoRS constructs a global state representation for $B^{\tau}$ via attention-based global pooling.

Both machine and operation nodes are aggregated using the same attention mechanism. Taking machines as an example, a shared linear scoring function $g(\cdot)$ is used to measure the importance of $M_k$ in the global representation, defined as $g_k=g(h_k^{\prime(\tau)})$, where $g_k\in\mathbb{R}$. The attention scores are normalized by softmax to obtain weights $\gamma_k$. The global machine representation is given by:
\begin{equation} 
h_{\text{global}}^{(m,\tau)}=\sum_{k=1}^{M}\gamma_k h_k^{\prime(\tau)}
\end{equation}
Similar to machines, operation nodes are aggregated by importance using the same attention pooling, giving $h_{\text{global}}^{(o,\tau)}=\sum_{i=1}^{n}\sum_{j=1}^{n_i}\gamma_{ij}h_{ij}^{\prime(\tau)}$, where $\gamma_{ij}$ is the attention weight of $O_{ij}$. The overall state of $B^{\tau}$ is constructed by concatenating the machine and operation global representations as $h_{\text{global}}^{\tau}=[h_{\text{global}}^{(m,\tau)}\|h_{\text{global}}^{(o,\tau)}]$, providing a global representation for decision-making and value evaluation.

\subsection{  Decoding Algorithm }
\label{subsec:state_reconstruction}

To reduce the inference overhead caused by frequent environment interactions and state updates, we propose a composite-interval decoding algorithm that minimizes these updates. At time step $t$, the feasible action set is defined as:
\begin{equation}
A_t=f_{\mathrm{mask}}\left(s_{t-1}^{\mathrm{feas}},a_{t-1}\right)
\end{equation}
 The function $f_{\text{mask}}(\cdot)$ updates the mask to restrict feasible actions without modifying state features. Specifically, $f_{\text{mask}}(\cdot)$ first updates the release times of the relevant machines according to the executed actions, and updates the current candidate operations of the scheduled jobs along with their machine compatibility relations. Then, based on the updated machine states, it calculates the earliest start time of each candidate operation--machine pair. Finally, operation--machine pairs that violate scheduling constraints are marked as infeasible, and the corresponding actions are masked out through the dynamic feasibility mask, thereby constructing the feasible action set for the current decision step. After completing $B^{\tau}$, the next input scheduling state representation 
$s^{(\tau+1)}$ is reconstructed once in a vectorized manner by 
$f_{\mathrm{recon}}(\cdot)$. Only the relative completion time of operations requires sequential updates due to strict order dependency, while other features are updated in batch by computing the effects of all actions in the composite interval. This produces the accurate state $s^{(\tau+1)}$ for subsequent scheduling decisions.

\subsection{Training Algorithm}

 CoRS is trained under an Actor–Critic framework~\citep{zhao2025dual}. Concatenated interval features are fed into the policy network, and the action distribution at state $s_t$ is defined as:
\begin{equation}
\pi_\omega(a_t\mid s_t)=\sigma\big(\text{MLP}_\omega[h_k^{\prime(\tau)} \| h_{ij}^{\prime(\tau)} \| h_{(ij,k)}^{(\tau)} \| h_{\text{global}}^\tau]\big)
\end{equation}
where $\sigma$ denotes softmax. The value network evaluates the current state and guides policy optimization, taking $h_{\text{global}}^\tau$ as input. The state value is defined as $v_\phi(s_t)=\text{MLP}_\phi(h_{\text{global}}^\tau)$. The policy is optimized using Proximal Policy Optimization with clipping~(PPO-Clip)~\citep{schulman2017proximal}, where the clipping mechanism constrains the magnitude of policy updates to improve training stability. Generalized Advantage Estimation~(GAE)~\citep{schulman2015high} is employed to compute the advantage function for improved optimization performance. The complete training procedure of CoRS is shown in Appendix C.

\section{Experiments}

This section describes the experimental setup, reports results on synthetic and public datasets, and analyzes the impact of the composite interval time step $X$ as well as the ablation results.

\subsection{Experimental Settings}
\textbf{Baselines.} The four selected PDRs are described as follows: FIFO schedules the OM pair that becomes available first; MOPNR schedules the OM pair whose operation has the largest number of remaining successor operations; SPT schedules the OM pair with the minimum processing time; MWKR schedules the OM pair with the maximum remaining average processing time. The three selected baselines are described as follows: HGNN~\citep{song2022flexible} models jobs and machines with a heterogeneous graph neural network to capture the relationships between operations and machines; DANIEL~\citep{wang2023flexible} adopts a reinforcement learning framework with a dual-attention network to improve solution quality; DOAGNN~\citep{zhao2025dual} decouples the disjunctive graph into a dual graph structure and applies a dual operation aggregation graph neural network  to represent operation states and their dependencies;ReSched~\citep{xiao2026resched} proposes a Transformer-based simplified state representation framework, improving solution quality.

\textbf{Datasets.}  We use $n \times m$ to denote an FJSP instance with $n$ jobs and $m$ machines. The model is trained on randomly generated instances of sizes $10 \times 5$, $20 \times 5$, $15 \times 10$, and $20 \times 10$, and evaluated on two synthetic datasets~(SD1, SD2) and four public datasets~(Hurink, Barnes, Brandimarte, and Dauzere). Detailed dataset descriptions are provided in Appendix D. Implementation details are in Appendix E.

\subsection{Experiment Results and Analysis}
\subsubsection{Performance Comparison}

Table 6~(Appendix F) and Table 7~(Appendix G) report the performance of CoRS and baseline methods on the synthetic and public datasets. LB denotes the known optimal solution for public datasets, while OR-Tools denotes results obtained by the Google OR-Tools exact solver on synthetic datasets with a time limit of 1800 s. Metrics include makespan~($C_{\max}$), relative gap to LB or OR-Tools~(Gap), and average solving time per instance~(Time/s). G and S denote the greedy and sampling strategies, respectively, where the sampling strategy performs 100 samples by default.

On the synthetic datasets, CoRS achieves significantly faster inference than all AR baselines. When $X=2$, the speed improves by 42.94\%--83.14\%; when $X=5$, the improvement further increases to 154.46\%--238.76\%. On the public benchmark datasets, CoRS  achieves a 160.14\%--489.86\% speedup at $X=5$. This acceleration mainly benefits from the CR mechanism, which reuses state encoding in a NAR manner within composite intervals, thereby substantially reducing state encoding operations.

While maintaining efficient inference, CoRS still produces high-quality solutions.  When $X=5$, the average gap to the best model on SD1 is only 4.04\%, which further reduces to 0.71\% on the public benchmarks. Compared with PDRs, CoRS also achieves an 18.27\%--41.18\% improvement with comparable solving time. These results mainly benefit from the enhanced static features. Although increasing $X$ reduces the model’s perception of dynamic information and degrades solution quality, the enhanced static features can still effectively represent the global state, thereby maintaining stable performance.

\subsubsection{Generalization Ability}

We train the model on $20 \times 10$ instances and evaluate its generalization on $30 \times 10$, $40 \times 10$, $60 \times 10$, and $100 \times 30$ instances, with results reported in Table 5~(Appendix H). When $X=3$, CoRS achieves high-quality solutions across all scales while maintaining the lowest inference time on SD1. 
On the 100$\times$30 instances, CoRS achieves a 8.96\% improvement in solution quality over AR models while maintaining lower inference time. 
On SD2, CoRS improves solution quality by 64.34\% over PDRs while achieving a 129.85\% inference speedup. These results demonstrate the strong generalization capability of CoRS. This is mainly due to the stable composite-interval state modeling of the interval encoding network and the effective constraints of the composite-interval decoding algorithm, allowing the model to maintain high inference efficiency and solution quality as the scale increases.

\subsubsection{Time-Constrained Performance Evaluation}

We evaluate solution quality under fixed average wall-clock budgets on the 15×10 instances from SD1. For each method and budget, we report the best Cmax obtained by completed rollouts or sampling settings whose average wall-clock solving time per instance does not exceed the budget. As shown in Table~\ref{tab:Performance under Fixed Time Budgets}, CoRS achieves competitive solution quality under tight time budgets. This advantage mainly stems from the CR mechanism, which reduces repeated state encoding and enables efficient schedule generation within limited inference time.

\begin{table}[]
\caption{Performance under Fixed Time Budgets}
\label{tab:Performance under Fixed Time Budgets}
\resizebox{\columnwidth}{!}{%
\begin{tabular}{lcccccc}
\hline
Time & DANIEL & RESHED & HGNN & DOAGNN & CoRS(X=2) & CoRS(X=5) \\
\hline
0.30 & -- & -- & -- & -- & -- & \textbf{163.39} \\
0.50 & -- & -- & -- & -- & \textbf{160.36} & 163.19 \\
0.80 & -- & -- & 174.47 & 171.61 & 160.36 & \textbf{159.27} \\
1.00 & 162.10 & 160.10 & 174.47 & 171.61 & \textbf{157.41} & 158.96 \\
1.50 & 160.05 & 160.06 & 170.96 & 168.05 & \textbf{156.67} & 157.88 \\
2.00 & 159.86 & 157.16 & 170.31 & 166.62 & \textbf{155.90} & 157.15 \\
4.00 & 157.38 & 156.88 & 166.63 & 164.05 & \textbf{154.63} & 155.64 \\
\hline
\end{tabular}%
}
\end{table}

\subsubsection{Impact of Composite Interval Steps~(X)}
We analyze the effect of $X$ on model performance using $10 \times 5$ instances from SD1, normalizing all metrics relative to the AR case~($X=1$). As shown in Figure 4~(Appendix J), increasing $X$ substantially reduces inference time, which is only 22.1\%--23\% of AR at $X=10$, while solution quality remains stable, varying by 3.3\%--7.6\%. This is primarily attributed to the proposed CR mechanism, which reduces inference time by decreasing state encoding, while the combination of enhanced static features with dynamic features maintains stable performance at larger time steps. In practical applications, the selection of $X$ should be adjusted according to the trade-off between inference efficiency and solution quality. For scenarios with stringent real-time requirements, a larger $X$ can be adopted to reduce the frequency of state encoding and  decrease inference latency. In contrast, for scenarios where solution quality is prioritized, a smaller $X$ should be selected to increase the frequency of state updates and maintain high-quality scheduling solutions.

\subsubsection{Ablation Studies}

As shown in Table 4~(Appendix I), we conduct a series of ablation experiments to evaluate the contributions of enhanced static features, the interval encoding method, and the composite-interval  decoding strategy. The results show that removing enhanced static features~(i.e., w/o En-fea) weakens node state representation, resulting in reduced solution quality. Furthermore, we analyze the interaction between $X$ and enhanced features, as shown in Table \ref{tab:x_feature}. The experimental results demonstrate that the performance gains brought by enhanced static features gradually increase with larger $X$. This indicates that enhanced static features provide  richer and more persistent structural information within composite intervals, thereby enabling stable decision-making. Removing the interval encoding method~(w/o Int-encoder) restores the step-by-step encoding strategy, resulting in only a slight improvement in average solution quality of 2.13\% while substantially increasing solving time by 223.81\%. Removing the composite-interval  decoding strategy~(i.e., w/o Co-decoder) causes more frequent state–environment interactions, thereby reducing inference efficiency. 

Furthermore, since CoRS is built upon the DANIEL framework, we conduct a 
progressive comparison from the original DANIEL to the complete CoRS model.  As shown in Table \ref{tab:daniel_cors}. 
Compared with  DANIEL, introducing interval state reuse alone
significantly reduces inference time, demonstrating that the CR mechanism effectively 
accelerates inference through interval-level state reuse. After further 
incorporating enhanced static features, CoRS  improves 
solution quality, indicating that enhanced static features effectively 
compensate for the performance degradation caused by the reduced state 
update frequency during interval state reuse.

\begin{table}[]
\caption{Interaction between $X$ and enhanced static features }
\label{tab:x_feature}
\centering

\begin{tabular}{cccccc}
\hline
\multirow{2}{*}{Method} & 
\multirow{2}{*}{} & 
\multicolumn{4}{c}{$X$} \\ 
\cline{3-6}

& & 2 & 3 & 4 & 5 \\ 
\hline

\multirow{2}{*}{CoRS}
& G & 107.91 & 108.26 & 108.93 & 109.35 \\
& S & 101.94 & 102.18 & 101.94 & 102.62 \\

\hline

\multirow{2}{*}{w/o En-fea}
& G & 108.89 & 110.25 & 111.66 & 113.56 \\
& S & 102.34 & 103.45 & 104.55 & 109.63 \\

\hline
\end{tabular}

\end{table}


\begin{table}[]
\caption{Progressive analysis from DANIEL to CoRS}
\label{tab:daniel_cors}
\resizebox{\columnwidth}{!}{%
\begin{tabular}{ccccccccc}
\hline
\multirow{2}{*}{Rdata} & \multicolumn{2}{c}{DANIEL} &  & \multicolumn{2}{c}{DANIEL+CR} &  & \multicolumn{2}{c}{CoRS} \\ \cline{2-3} \cline{5-6} \cline{8-9} 
 & Cmax & Time &  & Cmax & Time &  & Cmax & Time \\ \hline
G & 1031.95 & 0.912 &  & 1044.98 & 0.307 &  & 1032.45 & 0.213 \\
S & 978.33 & 3.25 &  & 991.05 & 1.118 &  & 982.23 & 0.895 \\ \hline
\end{tabular}%
}
\end{table}


\section{Conclusion}

This paper proposes the Composite Regressive (CR)  mechanism  and develops the CoRS scheduling model based on this  mechanism  for solving FJSP. CoRS adopts a two-timescale state evolution mechanism that decouples the interval-level representation state from the stepwise feasibility state. Furthermore, CoRS designs a feasibility-preserving composite-interval decoding scheme
, reducing state refresh overhead while maintaining sequential and feasible schedule construction.  In addition, enhanced static features are introduced to alleviate representation staleness caused by interval-level state reuse. Extensive experiments demonstrate that CoRS significantly reduces inference time while maintaining competitive solution quality, achieving superior speed--quality trade-offs under tight time budgets and low-latency scenarios. A current limitation is that the composite interval length is manually selected. The current method assumes a static deterministic FJSP and an exact stepwise feasibility tracker. Moreover, stale representations may still affect the ranking among feasible actions, and the achievable speedup depends on graph encoding
being a major component of inference cost. Future work will study adaptive interval selection and extend composite interval state reuse to stronger Transformer-based FJSP backbones for improved speed--quality trade-offs across instance scales and distributions.


\newpage

\bibliography{aaai2027}


\end{document}